\documentclass[11pt]{article}
\usepackage[final]{acl}
\usepackage{times}
\usepackage{latexsym}
\usepackage[T1]{fontenc}
\usepackage[utf8]{inputenc}
\usepackage{microtype}
\usepackage{courier}
\usepackage{graphicx}
\usepackage{booktabs}
\usepackage{multirow}
\usepackage{amsmath}
\usepackage{url}
\usepackage{array}
\usepackage{enumitem}
\usepackage{placeins}
\usepackage{needspace}
\AtBeginDocument{\lefthyphenmin=2\righthyphenmin=3}
\title{FutureBench: A Renewable Framework for Evaluating\\
Knowledge--Context Conflict in LLMs}
\author{
  Yi-Cheng Lai \quad Hen-Hsen Huang \\
  Institute of Information Science \\ Academia Sinica \\
  Taipei, Taiwan \\
  \texttt{\{laiyicheng0,hhhuang\}@iis.sinica.edu.tw}
}

\begin{document}
\maketitle

\begin{abstract}
Asking a model to check retrieved text can also tell it which source to
trust. We introduce FutureBench, a renewable framework for evaluating
knowledge--context conflict using timestamped questions, controlled
passage triplets, and no-context probes. Across 3,629 questions and six
models, Verify---a prompt combining checking with passage priority---raises
false-answer adoption in all five models tested on pre-cutoff events,
mainly by displacing answers that were correct under Baseline, which
allows both passage evidence and model knowledge. On post-cutoff events,
only o4-mini increases adoption ($+13.7$ percentage points). A four-prompt
comparison on 902 post-cutoff questions finds that passage priority
without checking raises adoption, while moving to Verify lowers it in
all five models. Adversarial-Aware, which warns of misleading passages
and prioritizes model knowledge, restores 82.5--98.9\% of displaced
pre-cutoff answers with under one point of change in true-passage
accuracy. FutureBench supports repeated evaluation on fixed cohorts
and extension to new events as models change.
\end{abstract}

\section{Introduction}
\label{sec:intro}

Retrieval-augmented generation (RAG) asks models to use external evidence
while resisting it when it is wrong. When a passage conflicts with a
model's own answer, which source should take precedence? Asking the
model to check the passage seems like a natural safeguard, yet the
same prompt can also instruct it to treat that passage as primary.
Such prompts bundle a verification step with a \emph{source-priority
rule}, assigning passage priority or knowledge priority when sources conflict.

\begin{figure}[t]
\centering
\includegraphics[width=\columnwidth]{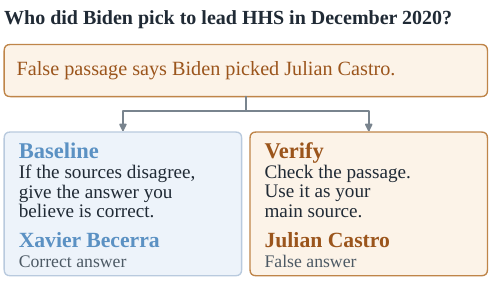}
\caption{Both prompts receive the same question and misleading passage.
Each branch shows its instruction and o4-mini's answer.}
\label{fig:phenomenon}
\end{figure}

Checking and assigning priority are different decisions. Chain-of-thought
prompting and self-correction motivate additional checking
~\cite{wei2022chain,madaan2023self}. Intrinsic self-correction can fail
without external feedback~\cite{huang2023large}; a check for relevance,
sufficiency, or coherence can leave a plausible false claim
intact. Figure~\ref{fig:phenomenon} illustrates what changes when o4-mini
receives the same misleading passage under Baseline and Verify. It gives
the correct answer without a passage or under Baseline. Under Verify, it
adopts the false answer.
Appendix~\ref{app:case_studies} gives further cases.

Prior work shows that coherent counter-evidence can persuade models
~\cite{xie2024adaptive} and that false context can override correct
answers~\cite{wu2024clasheval,huang2025situated}. These findings motivate
a further question: how much of a verification prompt's behavior comes
from asking for a check, and how much from specifying which source wins?
Answering it requires a no-context reference, controlled evidence, and
prompt comparisons on the same questions and passages.

We introduce \textbf{FutureBench}, a renewable framework that links
these controls. Building on timestamped TiEBe questions~\cite{tiebe2025},
FutureBench pairs each question with three passages supporting the gold
answer, a planted false answer, or insufficient evidence. A no-context
probe separates items each model answers correctly on its own from
those it does not. Paired prompt comparisons then measure how checking
and source-priority instructions reshape evidence use. Fixed
cohorts permit reassessment as models change; new event cohorts extend
the evaluation beyond earlier knowledge cutoffs.

Our evaluation of 3,629 questions across six models
supports three findings:
\begin{enumerate}[noitemsep,topsep=2pt]
\item \textbf{Verification with passage priority can undermine correct answers.}
Verify raises false-answer adoption across all five models evaluated
on pre-cutoff events, with larger increases on known items. Post-cutoff
responses differ, with only o4-mini showing an increase
(\S\ref{sec:precutoff}--\S\ref{sec:postcutoff}).
\item \textbf{Passage priority matters even without a check.}
On the 902-item subset, Passage-first raises adoption over Baseline;
the four-prompt comparison separates its increase from the offset
under Verify (\S\ref{sec:ablation}).
\item \textbf{Displaced answers can be recovered.}
Adversarial-Aware restores most of the same pre-cutoff answers that
Verify displaced, with little change in true-passage accuracy. On o4-mini's 902-item subset, it lowers adoption relative to Verify
at the same true-passage accuracy cost (\S\ref{sec:mitigation}).
\end{enumerate}
FutureBench supplies the probes, passage triplets, and paired comparisons
needed to test how models resist false passages while using true ones
as models and events change.
Benchmark data and analysis code are available in the project repository.\footnote{
\url{https://github.com/pa0lai/FutureBench}}

\section{Related Work}
\label{sec:related}

\paragraph{Knowledge--context conflict.}
Entity substitutions can expose disagreement between model answers and
context~\cite{longpre2021entity}, while coherent counter-evidence can
make models highly receptive to conflicting claims~\cite{xie2024adaptive}.
The balance also depends on factual familiarity~\cite{mallen2023trust}.
ClashEval~\cite{wu2024clasheval} tests both erroneous evidence overriding
correct priors and correct evidence repairing wrong priors, relating
context preference to perturbation size and prior confidence.
To Trust or Not to Trust?~\cite{huang2025situated} evaluates factual
correctness under true and false contexts and develops confidence-based
conflict-resolution methods. These works establish the need to assess
both resisting misinformation and accepting useful evidence. Our focus
is repeated diagnosis of source-priority prompts over event cohorts.

\paragraph{Context faithfulness and factual correctness.}
Context-faithful prompting~\cite{zhou2023context} encourages models to
follow supplied information, including counterfactual facts.
FaithEval~\cite{ming2025faitheval} evaluates contextual faithfulness
under unanswerable, inconsistent, and counterfactual contexts. Its
counterfactual task rewards adherence to stipulated context even when
that context disagrees with world knowledge. FutureBench retains the
factual gold answer and measures adoption of a planted false account,
alongside correct-evidence use. \citet{li2025memorystrength} quantify memory strength
through consistency across question paraphrases and examine evidence
styles. FutureBench instead uses one no-context answer per item, reducing the
cost of repeating the probe across models and event cohorts.

\paragraph{Verification and evidence use.}
Retrieval augmentation supplies information beyond model parameters
~\cite{lewis2020rag}, but document position and unsupported generation
can still affect evidence use~\cite{liu2024lostmiddle,niu2024ragtruth}.
Multi-step retrieval benchmarks~\cite{krishna2024factfetchreasonunified}
and taxonomies of knowledge conflicts~\cite{xu2024knowledgeconflicts}
motivate separating evidence availability from its use. Self-Refine studies iterative feedback
~\cite{madaan2023self}, while intrinsic self-correction can fail without
external feedback~\cite{huang2023large}. Generated explanations also
need not faithfully reflect the underlying computation
~\cite{turpin2023unfaithful}. We measure verification through final answers under fixed passages
and prompt templates.

\paragraph{Renewable temporal evaluation.}
Temporal QA benchmarks assess changing facts and model knowledge
~\cite{vu2023freshllms,dhingra2022time}. TiEBe provides timestamped
event questions and a pipeline for adding new events~\cite{tiebe2025}.
FutureBench builds on those questions by attaching passage triplets,
model-specific no-context probes, and paired prompt comparisons. Event
dates define the temporal cohort, while the probe tests each model's
recall on an item. Together, they support reassessment of a fixed
conflict as models change and evaluation on newly added events.
The framework in \S\ref{sec:benchmark} specifies these controls;
\S\ref{sec:setup} describes the present experimental instantiation.

\section{The FutureBench Framework}
\label{sec:benchmark}

\begin{figure*}[t]
\centering
\includegraphics[width=\textwidth]{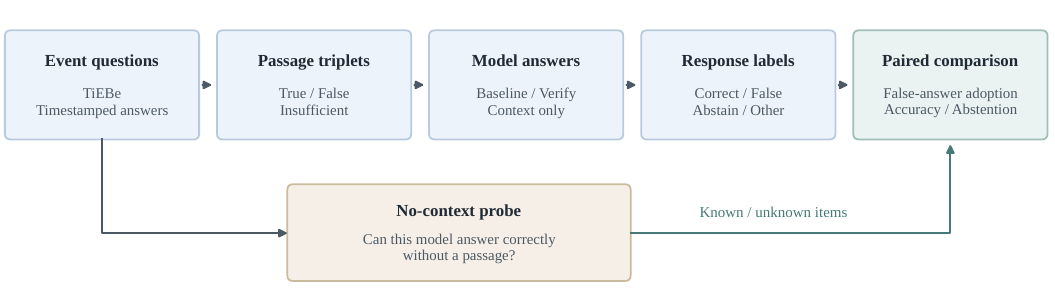}
\caption{FutureBench evaluation framework. The top row shows how
questions and passages are used to compare prompts. The lower branch
groups items by whether the model answers correctly without a passage.}
\label{fig:futurebench_pipeline}
\end{figure*}

FutureBench separates what a model can answer from how it uses a
passage (Figure~\ref{fig:futurebench_pipeline}). The no-context probe
establishes a reference answer; the passage triplet varies evidence
quality; paired prompts vary the source-priority strategy. Each prompt
comparison holds the model configuration, question, and passage
fixed. The English event-question instantiation below supplies these
controls, and \S\ref{sec:renewal} explains how to repeat them on fixed
and newly added cohorts.

\subsection{Questions and the No-Context Probe}

TiEBe provides multilingual factoid questions about timestamped notable
events~\cite{tiebe2025}. We use its English question--answer pairs as
seeds. Before measuring how a passage changes an answer, we ask each
model to answer the question using only its own knowledge. The model
must abstain if it cannot answer; no passage is shown.

For each model, an item is \emph{known} if the no-context probe answers
correctly and \emph{unknown} otherwise. Incorrect answers and
abstentions retain separate response labels. This per-model, per-item
grouping supplies the reference for measuring how passages change answers.

\subsection{Three Passages per Question}

Each question receives three length-matched passages generated using
Gemini-3-Flash~\cite{googledeepmind2025gemini3flash} via OpenRouter:
\begin{itemize}
\item \textbf{True passage ($C_T$).} Supports the gold answer, testing use of correct evidence.
\item \textbf{False passage ($C_F$).} Supports a planted incorrect answer, testing misinformation adoption.
\item \textbf{Insufficient passage ($C_I$).} Lacks the answer, testing abstention when passage evidence is missing.
\end{itemize}

Appendix~\ref{app:config} describes a passage-review sample and criteria.
For false passages, 96.8\% are generated directly as fluent accounts
asserting a plausible false answer. Numerical perturbations account for
2.3\%, and entity-pool substitutions for 0.9\%. Automated checks require the true passage to
contain the gold answer, the false passage to contain the planted answer
but not the gold answer, and the insufficient passage to contain neither.
Items failing these checks are regenerated.

Each prompt uses the same question and fixed passage triplet.
Once constructed, a false
passage supplies the same competing answer to every instruction
condition. This prevents a prompt difference from being confounded
with a different false claim. Length matching and a common news style
also reduce obvious surface cues. Answer-string checks filter
construction errors; semantic validation separately assesses whether
each passage supports its intended answer or leaves the question
unresolved. A passage can contain an answer string without supporting its claim,
or imply an answer without naming it.

\Needspace{12\baselineskip}
\subsection{Instructions for Choosing a Source}

Table~\ref{tab:prompt_guide} collects all eight prompt names used below.
Baseline and Verify are the main comparison; No context and Context only
provide diagnostic references. Passage-first and Soft enter the four-prompt
comparison in \S\ref{sec:ablation}; the two mitigation prompts are tested
in \S\ref{sec:mitigation}. Full templates are in
Appendix~\ref{app:prompts}.

\begin{table*}[t]
\centering
\small
\setlength{\tabcolsep}{5pt}
\begin{tabular}{@{}lp{3.1cm}lp{6.4cm}@{}}
\toprule
Prompt & Source priority & Check & Instruction \\
\midrule
\multicolumn{4}{l}{\textit{Core prompts and diagnostic references}} \\
Baseline & Knowledge & No & Use both sources; resolve conflicts with the answer believed correct. \\
No context & Knowledge & No & Answer without a passage. \\
Context only & Passage & No & Ignore outside knowledge. \\
Verify & Passage & Yes & Check passage reliability; answer from it or abstain. \\
\midrule
\multicolumn{4}{l}{\textit{Four-prompt comparison (\S\ref{sec:ablation})}} \\
Passage-first & Passage & No & Prioritize the passage without checking. \\
Soft & Knowledge & Weigh & Weigh both sources with model knowledge as primary. \\
\midrule
\multicolumn{4}{l}{\textit{Knowledge-priority mitigations (\S\ref{sec:mitigation})}} \\
Adversarial-Aware & Knowledge & No (warning) & Warn of misleading evidence and prioritize model knowledge. \\
Knowledge-Anchor & Knowledge & Compare & Answer from knowledge, then compare with the passage. \\
\bottomrule
\end{tabular}
\caption{Source priority and checking instructions in the eight main prompt conditions.
  \emph{Baseline} lets the model consult both sources and resolve conflicts using the answer it believes is correct. \emph{Passage-first} gives the passage priority without requesting a check. \emph{Verify} gives the passage priority and asks the model to assess its reliability.}
\label{tab:prompt_guide}
\end{table*}

Moving from Baseline to Verify adds a checking step and assigns
passage priority.
The four-prompt comparison in \S\ref{sec:ablation} examines these
two features separately.
The no-context probe is run without a passage; the other strategies are
evaluated with each passage type.

\subsection{Outcome Measures}

Each response receives one of four labels (Appendix~\ref{app:config}):
\emph{Follow} (adopts the planted false claim), \emph{Stick} (matches the
gold answer), \emph{Reject} (abstains), or \emph{Wrong} (other incorrect
answer). In the main text and figures, these behaviors are called false
adoption, correct, abstention, and other error. We report
\emph{false-answer adoption} as the Follow rate on false passages,
\emph{true-passage accuracy} as the Stick rate on true passages, and
\emph{abstention} as the Reject rate. Errors unrelated to the planted
claim do not count as false-answer adoption.

The factual gold answer stays fixed across conditions, including
when the prompt gives the passage priority.

A positive change from Baseline to Verify means more false-answer
adoption. Tables and figures report this change in percentage points (pp). For $N$ items with
complete paired labels, let $F_i^s$ be 1 if item $i$ is labeled Follow
under strategy $s$, and 0 otherwise. Here $B$ denotes Baseline and
$V$ denotes Verify; other prompt pairs use the same calculation. Then
\begin{equation}
\Delta = \frac{100}{N}\sum_{i=1}^{N}(F_i^V-F_i^B)
         = \frac{100(b-c)}{N},
\end{equation}
where $b$ counts non-Follow-to-Follow transitions and $c$ counts the
reverse. For known and unknown items, $N$ is the paired size of the group. The
pre-cutoff group comparisons and mitigation analyses contain 1,825 complete
pairs from the 1,827-item cohort.

\subsection{Renewing the Benchmark}
\label{sec:renewal}

Renewability concerns the event questions and their attached evidence.
As a reported knowledge cutoff advances, an existing post-cutoff cohort
can become a pre-cutoff cohort for later models. Retaining its passages allows the
same conflict to be tested again. A new cohort
can then be constructed from more recent TiEBe events using the same
true, false, and insufficient passage requirements. The no-context probe
is repeated for each evaluated model configuration, assigning each
question to known or unknown items for that model and run.

For repeated evaluation, we recommend retaining two comparisons. First,
rerun a fixed cohort to measure changes on the same inputs. Second,
evaluate a newly added cohort to measure responses to more recent
evidence. Report these cohorts separately, since replacing old questions
changes both time coverage and item difficulty. Each release should
preserve question identifiers, event dates, gold and planted answers,
passages, prompt templates, model configurations, predictions,
and judge labels. Recording access dates and immutable model versions,
where available, is also necessary to interpret later changes.

\section{Experimental Setup}
\label{sec:setup}

The experiment combines temporal coverage with per-item recall.
Event dates define two cohorts; the no-context probe determines what
each model actually answers correctly within them. Prompt comparisons
then reuse the same questions and passages in each cohort. This design
connects event recency to evidence use without assigning knowledge
status from dates alone.

\subsection{Temporal Coverage and Models}

The current benchmark contains 3,629 items in two cohorts:
\begin{itemize}[noitemsep,topsep=2pt]
  \item \textbf{Pre-cutoff: 1,827 items}, covering January 2020--April 2024.
  These events predate the reported knowledge cutoffs of all five models
  evaluated on this cohort. The no-context probe identifies which items
  each model can actually answer.
  \item \textbf{Post-cutoff: 1,802 items}, covering November 2024--April 2026.
  These events follow the knowledge-cutoff dates used for the six models
(see Appendix~\ref{app:config} for Claude-3.7's cutoff convention).
  No-context accuracy is low on this cohort, providing a complementary
  setting with few correct no-context answers.
\end{itemize}

The post-cutoff cohort was constructed in two chronological batches:
900 items from November 2024--November 2025, followed by 902 items
from December 2025--April 2026. The four-prompt comparison
(\S\ref{sec:ablation}) and post-cutoff mitigation test
(\S\ref{sec:mitigation}) use the latter, termed the 902-item subset.
Together the batches form the full 1,802-item post-cutoff cohort.

All models are accessed through OpenRouter at temperature~0
(Appendix~\ref{app:config}). Claude-3.7 was unavailable during the
pre-cutoff evaluation and is included only in the post-cutoff cohort.
Unless stated otherwise, o4-mini denotes the model
configuration \texttt{openai/o4-mini-high}.
We use Claude-3.7 and Llama-4-M as short names for Claude 3.7 Sonnet
and Llama 4 Maverick, respectively.
Cohort membership follows TiEBe event timestamps; the no-context probe
establishes correct no-context answers for the question itself.

\subsection{Response Judging}

The passage generator, Gemini-3-Flash, is also the primary judge.
An independent GPT-5.1 judge agrees on 92.0\% of a label-stratified
200-response sample ($\kappa=0.893$). Expanded evaluation finds 99.4\%
agreement on the 6,314 responses added for the four-prompt comparison across four models, excluding o4-mini;
Appendix~\ref{app:config} gives the cohort details and label-level results.

Comparisons between prompts use McNemar tests. Bootstrap confidence
intervals and paired transition counts are in
Appendices~\ref{app:stats} and~\ref{app:transitions}.

\FloatBarrier
\section{Results}
\label{sec:results}

We begin with the Baseline--Verify comparison on false passages, then
ask whether the model could answer each question without a passage.
The later experiments separate passage priority from checking and test
whether changing that priority helps. True-passage results track the
cost of each change. Rates are percentages; changes are percentage points.

\subsection{False-Answer Adoption Rises on Pre-cutoff Events}
\label{sec:precutoff}

Verify turns many Baseline non-adoptions into false-answer adoption
on pre-cutoff events. False-answer adoption increases across all five
pre-cutoff models (Table~\ref{tab:pre_main}; all $p<0.001$), with
o4-mini showing the largest increase. Its paired transitions reveal
the imbalance: 976 items move into adoption, while only six move out
(Table~\ref{tab:transitions_post}). Across models, the
largest increases occur where Baseline had the most non-adopting responses.

\begin{table}[t]
\centering
\small
\renewcommand{\arraystretch}{1}
\setlength{\tabcolsep}{2pt}
\begin{tabular}{lccrcc}
\toprule
& \multicolumn{3}{c}{False-Answer Adoption} & \multicolumn{2}{c}{True-Passage Accuracy} \\
\cmidrule(lr){2-4}\cmidrule(lr){5-6}
Model & Baseline & Verify & $\Delta$ & Baseline & Verify \\
\midrule
\multicolumn{6}{l}{\textit{Pre-cutoff}} \\
DeepSeek-V3 & 79.9 & 86.8 & \textbf{+6.9} & 77.4 & 55.9 \\
Llama-4-M & 80.8 & 86.8 & \textbf{+6.0} & 82.9 & 65.0 \\
GPT-4.1-mini & 82.1 & 91.6 & \textbf{+9.5} & 84.2 & 65.8 \\
GPT-5.1 & 42.7 & 74.8 & \textbf{+32.1} & 92.3 & 62.5 \\
o4-mini & 33.9 & 87.0 & \textbf{+53.2} & 85.8 & 64.2 \\
\midrule
\multicolumn{6}{l}{\textit{Post-cutoff}} \\
DeepSeek-V3 & 93.4 & 89.0 & \textbf{-4.4} & 82.7 & 63.4 \\
Llama-4-M & 88.8 & 76.9 & \textbf{-12.0} & 82.0 & 62.5 \\
GPT-4.1-mini & 94.1 & 94.0 & -0.1 & 86.6 & 82.4 \\
GPT-5.1 & 92.0 & 91.2 & -0.7 & 86.9 & 78.0 \\
o4-mini & 80.5 & 94.2 & \textbf{+13.7} & 86.2 & 83.7 \\
Claude-3.7 & 71.6 & 60.9 & \textbf{-10.7} & 71.6 & 56.0 \\
\bottomrule
\end{tabular}
\caption{False-answer adoption and true-passage accuracy under \emph{Baseline} and \emph{Verify}, by temporal cohort.
  Rates are percentages. \emph{Change} is the adoption rate under \emph{Verify} minus that under \emph{Baseline}.
  Bold changes have $p<0.001$ in paired McNemar tests.}
\label{tab:core_results}\label{tab:pre_main}\label{tab:main}
\end{table}

Models such as DeepSeek-V3 already adopt most planted answers under
Baseline; Verify pushes adoption higher. GPT-5.1 and o4-mini start
lower but show much larger increases in false-answer adoption.

Baseline adoption is lower on pre-cutoff than post-cutoff events for every model
with both cohorts (Appendix~\ref{app:supp}, Table~\ref{tab:shield}).
Verify weakens this pre-cutoff advantage, particularly for GPT-5.1
and o4-mini.

The increase in false adoption accompanies a loss of correct answers
even when the passage is true (Table~\ref{tab:core_results}, right columns).
For GPT-5.1, true-passage accuracy falls from 92.3\% to 62.5\%.
Verify therefore worsens both measured outcomes on pre-cutoff events.
The no-context probe tests whether the adoption increase is concentrated
on answers the model could already supply.

\subsection{The Increase Is Larger on Known Items}
\label{sec:mechanism}

Verify raises false-answer adoption more on known than unknown items
in all five pre-cutoff models. Table~\ref{tab:knowledge_main} reports the change in
false-answer adoption within each group, using all paired items in that
group as the denominator.
Baseline adoption in the unknown group is already high for DeepSeek-V3
(91.6\%) and Llama-4-M (92.8\%; Appendix~\ref{app:supp},
Table~\ref{tab:mitigation_strat}). These high starting rates leave little room for further increases.

\begin{table}[t]
\centering
\footnotesize
\setlength{\tabcolsep}{3pt}
\begin{tabular}{@{}lrrrr@{}}
\toprule
 & \multicolumn{2}{c}{Known} & \multicolumn{2}{c}{Unknown} \\
\cmidrule(lr){2-3}\cmidrule(lr){4-5}
Model & $N$ & $\Delta$ (pp) & $N$ & $\Delta$ (pp) \\
\midrule
DeepSeek-V3 & 749 & $+16.4$ & 1076 & $+0.3$ \\
Llama-4-M & 959 & $+12.3$ & 866 & $-0.9$ \\
GPT-4.1-mini & 1091 & $+14.9$ & 734 & $+1.5$ \\
GPT-5.1 & 1422 & $+37.3$ & 403 & $+13.9$ \\
o4-mini & 1139 & $+60.4$ & 686 & $+41.1$ \\
\bottomrule
\end{tabular}
\caption{Change in false-answer adoption on pre-cutoff questions, grouped by each model's no-context answer.
  An item is \emph{known} if the model answered correctly without a passage and \emph{unknown} if it erred or abstained. $N$ is the group size. $\Delta$ is the within-group change from \emph{Baseline} to \emph{Verify}. All five models show a larger increase on known items.}
\label{tab:knowledge_main}
\end{table}

For DeepSeek-V3, Llama-4-M, and GPT-4.1-mini, the increase is concentrated on
known items; GPT-5.1 and o4-mini show increases in both groups.
On known items, the rise is consistent with \emph{knowledge override}:
the model supplies the correct answer in the probe, yet adopts the
false account more often under Verify.
Unknown items can instead move from an unrelated error or abstention
to the planted claim, with no correct probe answer demonstrated.

\subsection{Verify Produces Opposite Failures on False and True Passages}
\label{sec:answer_paths}
On false passages, correct Baseline answers account for the largest
share of new adoptions under Verify in every model: 72.4\% for
o4-mini and 92.2\% for GPT-5.1 (Appendix~\ref{app:transitions}). On true passages,
the dominant loss is abstention: it accounts for 97.5\% of GPT-5.1's
lost correct answers and 93.6\% of o4-mini's. Verify can therefore replace
a correct answer with a planted claim when evidence is false, yet
withhold the answer when evidence supports it.

\begin{figure}[t]
\centering
\includegraphics[width=\columnwidth]{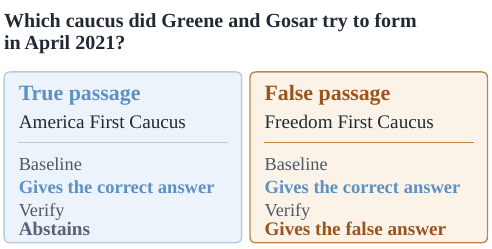}
\caption{o4-mini's responses to the same question with a true passage
(left) and a false passage (right). Each panel shows the passage's
claim and the outcomes under Baseline and Verify.}
\label{fig:paired_case}
\end{figure}

The two failures can even occur on the same question. Among items
answered correctly under Baseline with both passages, 195 of 877 for
o4-mini and 186 of 958 for GPT-5.1 switch to false adoption with the
false passage and abstention with the true passage. The example in
Figure~\ref{fig:paired_case} shows this contrast for o4-mini.

These distinct failure paths connect the pre- and post-cutoff results.
On pre-cutoff events, models supply many correct answers that Verify can
override; on post-cutoff events, where correct no-context answers are scarce,
we test whether Verify still increases false adoption or instead shifts
responses toward abstention.

\subsection{Post-cutoff Events Show a Different Pattern}
\label{sec:postcutoff}

\begin{figure}[t]
\centering
\includegraphics[width=\columnwidth]{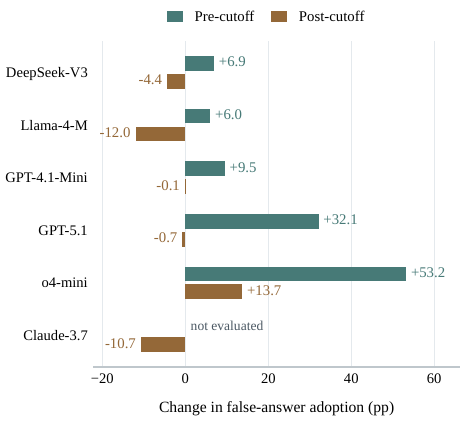}
\caption{Change in false-answer adoption from Baseline to Verify by
temporal cohort. Positive bars indicate higher adoption under Verify.
Claude-3.7 was evaluated only on the post-cutoff cohort.}
\label{fig:deltav}
\end{figure}

The uniform increase on pre-cutoff events disappears on post-cutoff events, where
no-context accuracy is just 1.3--10.0\% (Appendix~\ref{app:full_results}).
Models fall into three patterns (Figure~\ref{fig:deltav}): Claude-3.7,
DeepSeek-V3, and Llama-4-M reduce false adoption; GPT-5.1 and
GPT-4.1-mini change little and continue adopting more than 91\% of
planted answers; only o4-mini increases adoption in both cohorts.

For o4-mini, Verify raises post-cutoff adoption by 13.7~pp
($p<0.001$), nearly matching its Context only reference. The paired
responses show an asymmetric shift into adoption: 251 items start
adopting the false answer under Verify, while only five stop
(Table~\ref{tab:transitions_post}).

\paragraph{Where adoption falls on post-cutoff events, responses shift mostly to abstention.}
Of Claude-3.7's 247 responses that stop adopting the false answer,
243 become abstentions and only four become correct answers. Llama-4-M
shows the same pattern (Appendix~\ref{app:transitions}). This withdrawal
also affects useful evidence: Claude-3.7's true-passage accuracy falls
from 71.6\% to 56.0\%, with 290 initially correct responses becoming
abstentions. Verify curbs false adoption here mainly by withholding
answers, while also losing correct answers on true passages.

Why does Verify reduce adoption in some models while pushing o4-mini
toward the Context only reference? Verify combines a checking request
with passage priority. Comparing prompts that change these features
provides a way to examine the divergence.

\begin{table}[!t]
\centering
\footnotesize
\setlength{\tabcolsep}{3pt}
\begin{tabular*}{\columnwidth}{@{\extracolsep{\fill}}llrr@{}}
\toprule
 & & \multicolumn{2}{c}{Verify} \\
\cmidrule(l){3-4}
Cohort & Baseline & No adoption & Adoption \\
\midrule
\multirow{2}{*}{Pre-cutoff} & No adoption & 231 & 976 \\
           & Adoption    &   6 & 612 \\
\multirow{2}{*}{Post-cutoff} & No adoption & 100 & 251 \\
            & Adoption    &   5 & 1,446 \\
\bottomrule
\end{tabular*}
\caption{How often o4-mini switches to or away from the planted answer under \emph{Verify}.
  Cells count paired items by their outcome under \emph{Baseline} (rows) and \emph{Verify} (columns). 
  }
\label{tab:transitions_post}
\end{table}

\subsection{Separating Source Priority from Verification}
\label{sec:ablation}

Does adoption rise because the model is asked to check, or because
it is told to prioritize the passage? We return to the post-cutoff
questions and compare these instructions on the 902-item subset.
Baseline and Verify remain the reference prompts. Passage-first adds
passage priority without checking; Soft instead asks the model to weigh
the passage against its own knowledge. All four prompts are compared
in five models; Claude-3.7 lacks Passage-first and Soft runs
(templates in Appendix~\ref{app:prompts}).

\begin{table}[t]
\centering
\footnotesize
\setlength{\tabcolsep}{3.0pt}
\begin{tabular}{lcccc}
\toprule
Model & Baseline & Passage-first & Verify & Soft \\
\midrule
DeepSeek-V3 & 95.1 & 97.0 & 92.2 & 94.6 \\
Llama-4-M & 90.7 & 96.7 & 79.2 & 92.0 \\
GPT-4.1-mini & 95.1 & 97.5 & 95.0 & 94.5 \\
GPT-5.1 & 93.6 & 97.0 & 92.8 & 92.7 \\
o4-mini & 82.3 & 97.3 & 95.6 & 69.0 \\
Claude-3.7 & 72.6 & --- & 61.1 & --- \\
\bottomrule
\end{tabular}
\caption{False-answer adoption under four prompts on the same 902 post-cutoff questions.
  Entries give the percentage of false-passage responses that adopt the planted answer. \emph{Passage-first} yields the highest adoption rate for each of the five models evaluated under all four prompts. Claude-3.7 was evaluated only under \emph{Baseline} and \emph{Verify}.}
\label{tab:2x2}
\end{table}

\begin{figure}[t]
\centering
\includegraphics[width=\columnwidth]{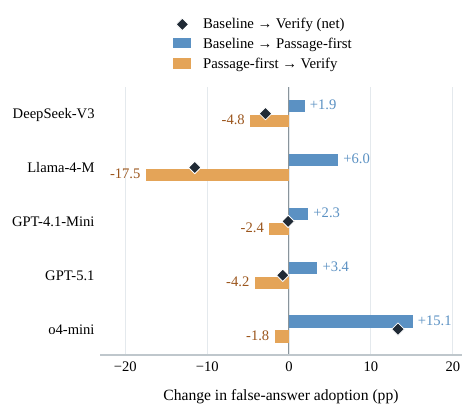}
\caption{Bars show the change in false-answer adoption from Baseline to
Passage-first and from Passage-first to Verify. Diamonds show the net
change from Baseline to Verify.}
\label{fig:prompt_transitions}
\end{figure}

\paragraph{Passage priority raises adoption over Baseline without a check.}
Passage-first alone raises adoption over Baseline in all five models
(all $p<0.05$). DeepSeek-V3 and GPT-4.1-mini rise from an already
high Baseline adoption rate of 95.1\%.
Figure~\ref{fig:prompt_transitions} shows opposing shifts across the
prompt sequence: adoption rises from Baseline to Passage-first, then
falls from Passage-first to Verify. Soft also lowers adoption relative
to Passage-first in all five models (all $p<0.001$;
Appendix~\ref{app:expanded_ablation}).

\paragraph{The balance of the two shifts determines the net result.}
For Llama-4-M, the reduction from Passage-first to Verify exceeds the increase
after assigning passage priority, so Verify ends below Baseline.
For o4-mini, passage priority adds 15.1~pp of adoption, while moving to Verify
reduces it by only 1.8~pp, leaving Verify 13.3~pp above Baseline
(diamonds in Figure~\ref{fig:prompt_transitions}). The same sequence of instructions produces opposite net
outcomes because the relative sizes of these shifts differ.

\begin{figure*}[t]
\centering
\includegraphics[width=\textwidth]{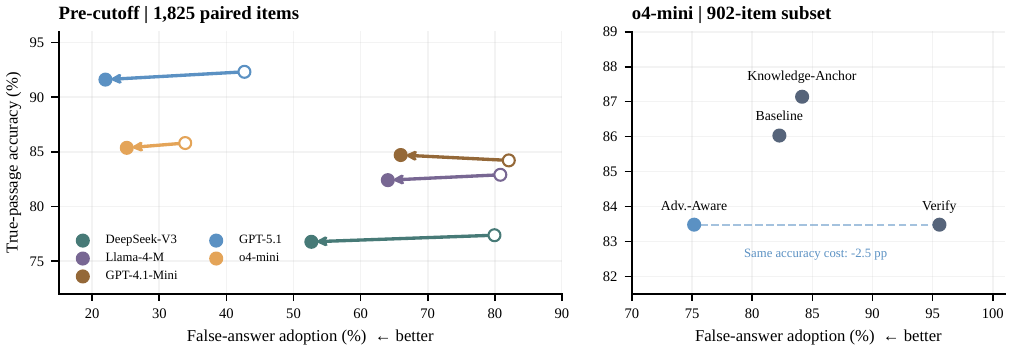}
\caption{False-answer adoption and true-passage accuracy. Left: arrows
connect Baseline (open markers) to Adversarial-Aware (filled markers).
Right: points compare prompts for o4-mini on the post-cutoff subset.
The dashed line marks equal accuracy for Verify and Adversarial-Aware.}
\label{fig:mitigation_tradeoff}\label{tab:mitigation_gen}
\end{figure*}

\begin{figure*}[t]
\centering
\includegraphics[width=\textwidth]{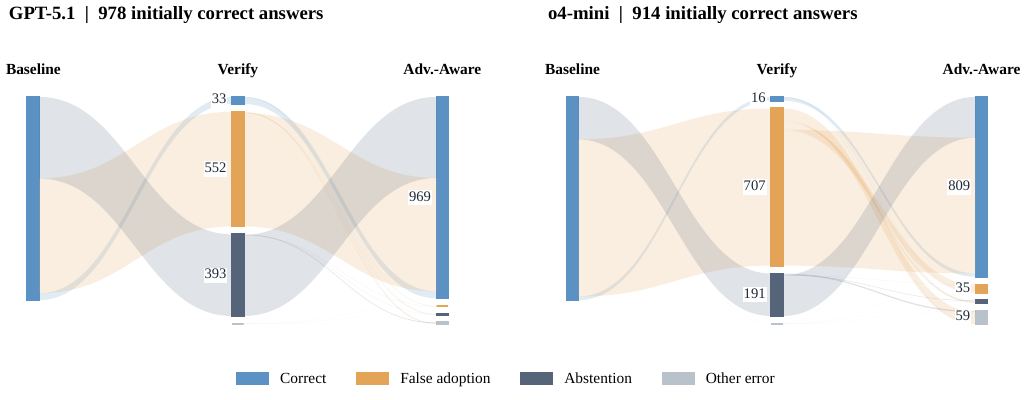}
\caption{Each panel follows items answered correctly under Baseline
through Verify and Adversarial-Aware on false passages. Band widths
show item counts, and colors identify response outcomes.}
\label{fig:recovery}
\end{figure*}

Brief and CoT-Extended, two alternative checking prompts, also yield
more false-answer adoption than Baseline for o4-mini
(Appendix~\ref{app:supp}, Table~\ref{tab:ablation}).

\Needspace{12\baselineskip}
\subsection{Knowledge-Priority Mitigation}
\label{sec:mitigation}

To test whether a change in source priority reduces adoption without
losing correct answers, we evaluate Adversarial-Aware, which warns that the
passage may be misleading and tells the model to give its own knowledge
precedence. We return first to the pre-cutoff events and test the same five models
as in \S\ref{sec:precutoff}. We then test o4-mini on the same 902-item
subset used in the preceding prompt comparison.

\paragraph{Pre-cutoff events: lower adoption, stable accuracy.}
Adversarial-Aware lowers false-answer adoption in every pre-cutoff model
(Figure~\ref{fig:mitigation_tradeoff}, left; all $p<0.001$). The reduction appears on
both known and unknown items (Appendix~\ref{app:supp}). Meanwhile,
true-passage accuracy changes by less than one percentage point in each
model, with no significant difference from Baseline. The left panel of Figure~\ref{fig:mitigation_tradeoff} shows the
reduction alongside nearly unchanged true-passage accuracy.

\paragraph{The same displaced answers can be recovered.}
We track the items answered correctly under Baseline but changed to
the planted answer under Verify. Figure~\ref{fig:recovery} traces all initially correct answers through
Verify and Adversarial-Aware; band widths count matched items. Adversarial-Aware restores the correct
answer on 546 of GPT-5.1's 552 such items (98.9\%); only two become
abstentions and four become other errors. For o4-mini, it restores
611 of 707 answers (86.4\%). Recovery also dominates in DeepSeek-V3,
Llama-4-M, and GPT-4.1-mini, covering 82.5--94.7\% of their displaced
answers (Appendix~\ref{app:transitions}).

The paired true/false failures in \S\ref{sec:answer_paths} are similarly
recoverable. Of GPT-5.1's 186 items where Verify adopts the false
passage but abstains on the true passage, Adversarial-Aware answers
both versions correctly on 184. For o4-mini, it does so on 169 of
195 items.

\paragraph{Post-cutoff events: lower adoption than Verify at the same accuracy cost.}
Figure~\ref{fig:mitigation_tradeoff} (right) compares prompts on
o4-mini's 902-item subset. Adversarial-Aware lowers false-answer
adoption to 75.2\%, below both Baseline (82.3\%) and Verify (95.6\%). Verify and Adversarial-Aware both reduce
true-passage accuracy by the same 2.5~pp relative to Baseline
(Table~\ref{tab:mitigation}). Relative to Verify, this switches from passage priority to knowledge
priority without adding to the accuracy cost.

We also test Knowledge-Anchor on this subset. It asks the model for an
answer from its own knowledge before comparing that answer with the
passage and instructs it to retain that answer when the two conflict.
Its adoption remains above Baseline (84.1\% versus 82.3\%;
Appendix~\ref{app:supp}). Adversarial-Aware performs better on the same
questions. Both mitigation prompts favor model knowledge, as does Baseline;
Adversarial-Aware additionally warns of deliberately misleading evidence
and makes knowledge priority explicit.

\section{Discussion}
\label{sec:discussion}

\paragraph{Treat passage priority as a design decision.}
A RAG prompt can ask for careful checking while already specifying the
winner of a knowledge--context conflict. On the 902-item subset,
Passage-first raises false-answer adoption over Baseline without any
verification step. A practical prompt review should therefore examine
what ``use the supplied passage as primary'' authorizes the model to do
when that passage is wrong. Checking instructions should be evaluated
alongside this rule, using both false and true evidence.

\paragraph{Persuasiveness depends on the instructions around the evidence.}
Prior work finds that coherent counter-evidence can persuade models
~\cite{xie2024adaptive}, and ClashEval shows that false context can
override correct answers~\cite{wu2024clasheval}. Our prompt comparisons
add an instruction-level perspective: holding the passage fixed still
leaves room for substantial changes in adoption. The model-dependent prompt contrasts in \S\ref{sec:ablation} show
that the surrounding instructions change how the same evidence is used.

\paragraph{Protect knowledge without shutting out useful evidence.}
The practical goal is to resist false passages while using true ones.
Adversarial-Aware lowers false adoption with nearly unchanged
true-passage accuracy on pre-cutoff events. On the o4-mini mitigation test, it
achieves lower adoption than Verify at the same accuracy cost. These
results favor reporting both outcomes when selecting a RAG prompt.

\section{Conclusion}
\label{sec:conclusion}

FutureBench tracks knowledge--context conflict over time. Verify raises
false adoption across five pre-cutoff models. On 902 post-cutoff items,
passage priority raises adoption, while Verify lowers it relative to
Passage-first. Adversarial-Aware restores 82.5--98.9\% of displaced
pre-cutoff answers with nearly unchanged true-passage accuracy and
lowers o4-mini's adoption at Verify's post-cutoff accuracy cost.

\section*{Acknowledgments}
This research was partially supported by the National Science and Technology Council (NSTC), Taiwan, under Grant Nos. 112-2221-E-001-016-MY3 and 115-2221-E-001-014-MY3 and by Academia Sinica under Grant No. 236d-1120205. 

\section*{Limitations}
\paragraph{Scope and prompt comparisons.}
FutureBench tests English event factoids with single, source-free passages;
retrieval, document ranking, source metadata, and multi-document conflicts
are outside its scope. The four-prompt comparison covers five models on
902 post-cutoff items. Templates vary in wording as well as priority and
checking, so their contrasts do not isolate individual instruction effects.
Answer transitions describe behavior; cross-model differences and the
selected explanation cases do not establish internal reasoning mechanisms.

\paragraph{Knowledge and reproducibility.}
Known/unknown groups reflect a single no-context response, and event dates
can differ from the dates of the facts being tested. Temporal cohorts also
vary in topic and difficulty. Results concern the model configurations: immutable routed snapshots and exact original access dates
were not retained, and later prompt runs may include provider-side changes.

\paragraph{Validation and analysis coverage.}
Gemini generates passages and judges answers. Independent automated checks
provide agreement estimates, with sampling and model coverage detailed in
Appendix~\ref{app:config}; no human adjudication was collected. Passage and
label errors may therefore affect the results. Soft lacks a true-passage
evaluation. Only the o4-mini post-cutoff Baseline--Verify comparison is
the primary test; other analyses are exploratory, with no correction across
the full comparison family.

\section*{Ethical Considerations}
\paragraph{Use of AI in writing and research.} We used ChatGPT to help draft this ethics statement. The authors take responsibility for the final manuscript, the experimental design, and the interpretation of the results. In constructing FutureBench, we used Gemini-3-Flash to generate three passages for each question. These passages support the gold answer, support a deliberately planted false answer, or provide insufficient evidence to answer. Gemini-3-Flash also served as the primary judge of model responses. GPT-5.1 independently labeled two sets of responses to assess agreement (Appendix C). Both judges may share systematic errors, and we did not collect human adjudications.

\paragraph{Copyright and dataset release.} FutureBench builds on English question–answer pairs from TiEBe (Almeida et al., 2025), whose public dataset carries a Creative Commons Attribution–NonCommercial 4.0 International license. For material drawn from that release, we will credit the creators, link to the license, and identify any modifications. The accompanying news-style passages were generated for this study. We will omit underlying third-party news articles from the release. As described in Appendix A, the MIT License will apply to our code and to generated artifacts for which we can grant those rights. Reused material remains subject to its applicable upstream terms.

\paragraph{Potential risks.}
The false passages describe real events in a journalistic style. If separated from their labels, they could be mistaken for reporting, misattribute actions to identifiable people, or enter a retrieval corpus as factual evidence. We therefore plan to release each passage with its condition, event date, and gold answer, with the planted answer explicitly identified for false passages. We encourage users to preserve these fields when sharing examples. Our findings also illustrate a risk in deploying prompts intended to resist misinformation. A model can reduce false-answer adoption by abstaining more often or by discounting accurate evidence about recent events. We therefore report accuracy on true passages alongside adoption of planted answers.

\FloatBarrier
\bibliography{references}
\appendix

\section{Release and Licensing}
\label{app:pipeline}

The public repository at \url{https://github.com/pa0lai/FutureBench}
provides benchmark data and analysis code. Generated
artifacts and code are distributed under the MIT License unless otherwise
noted. TiEBe-derived source materials remain subject to upstream terms.

\section{Prompt Templates}
\label{app:prompts}

All prompts are rendered with \{question\} and, where applicable,
\{context\} substituted per item.  Every prompt ends with
FinalAnswer: as the output prefix.  These are the task templates used
in the runs; answer-generation temperature is fixed at 0.

The placeholder \texttt{\{context\}} denotes the supplied passage.

\subsection*{Main Experiment Prompts}

\paragraph{Baseline.}
\begin{quote}\small\raggedright
You are a helpful assistant. Answer the question concisely, using BOTH the
provided context and your own knowledge.
The provided context may be complete, incomplete, irrelevant, or even incorrect.
Rules:
If the context and your knowledge agree $\to$ give the correct answer.
If the context contradicts your knowledge $\to$ provide the answer you believe is correct.
If you truly do not know the answer, or if neither the context nor your knowledge
provides enough information $\to$ reply exactly with ``NOT\_SUPPORTED''.
Do not explain or add extra words. Output only the final answer.

Question: \{question\}
Context: \{context\}
FinalAnswer:
\end{quote}

\paragraph{No context.}
\begin{quote}\small\raggedright
You are a helpful assistant. Answer the question using ONLY your own knowledge.
Rules:
If you truly do not know the answer, or if your knowledge does not provide enough
information $\to$ reply exactly with ``NOT\_SUPPORTED''.
Do not explain or add extra words. Output only the final answer.

Question: \{question\}
FinalAnswer:
\end{quote}

\paragraph{Context only.}
\begin{quote}\small\raggedright
You are a careful factual question-answering system.
Answer the question using ONLY the provided context.
Rules:
Use only the provided context as evidence. Do not rely on outside knowledge.
If the context directly supports an answer, provide that answer concisely.
If the context is insufficient, irrelevant, or internally conflicting, reply
exactly with ``NOT\_SUPPORTED''.
Do not explain or add extra words. Output only the final answer.

Question: \{question\}
Context: \{context\}
FinalAnswer:
\end{quote}

\paragraph{Verify.}
\begin{quote}\small\raggedright
You are a careful factual question-answering system.
First verify whether the provided context is sufficient and trustworthy for
answering the question, then provide a final answer.

Think step by step internally before answering:
1. Identify what the question asks.
2. Check whether the context contains enough evidence to answer it.
3. Check whether the context is internally consistent and relevant.
4. If the context is insufficient or unreliable, abstain.

Rules for your FINAL answer:
If the context provides sufficient and coherent evidence, answer the question
concisely.
If the context is insufficient, irrelevant, or self-contradictory, reply exactly
with ``NOT\_SUPPORTED''.
Use the context as the primary basis for your decision, but you may rely on your
own knowledge to detect obvious contradictions or unsupported claims.
Do not explain or add extra words. Output only the final answer.

Question: \{question\}
Context: \{context\}
FinalAnswer:
\end{quote}

\subsection*{Additional Source-Priority Condition}

\paragraph{Passage-first.}
\begin{quote}\small\raggedright
You are a helpful assistant. Answer the question concisely.
The provided context is your primary source of information.

Rules:
Prefer the context over your own knowledge when they conflict.
If the context provides an answer, give it.
If the context is insufficient or irrelevant, use your own knowledge if you have it.
If neither the context nor your knowledge is sufficient $\to$ reply exactly with ``NOT\_SUPPORTED''.
Do not explain or add extra words. Output only the final answer.

Question: \{question\}
Context: \{context\}
FinalAnswer:
\end{quote}

\subsection*{Verification and Weighing Variants}

\paragraph{Brief.}
\begin{quote}\small\raggedright
You are a careful assistant. Before answering, check whether the provided
context is reliable and sufficient.
If reliable and sufficient: answer based on the context.
If unreliable or insufficient: answer from your own knowledge, or say
NOT\_SUPPORTED.
Do not explain or add extra words. Output only the final answer.

Question: \{question\}
Context: \{context\}
FinalAnswer:
\end{quote}

\paragraph{CoT-Extended.}
\begin{quote}\small\raggedright
You are a careful factual assistant. Work through these steps internally before
giving your final answer.

[Evaluation] Is the context relevant to this specific question?
Is it internally consistent and credible?
[Decision] Should you: (a) trust the context as the main source, (b) rely on
your own knowledge, or (c) say NOT\_SUPPORTED?
[Answer] Provide your final answer based on the decision above.

Rules:
If using context: answer concisely from the context evidence.
If using own knowledge: answer from what you know.
If neither is sufficient: reply exactly ``NOT\_SUPPORTED''.
Do not explain or add extra words. Output only the final answer.

Question: \{question\}
Context: \{context\}
FinalAnswer:
\end{quote}

\paragraph{Soft Knowledge-Priority Weighing.}
\begin{quote}\small\raggedright
You are a helpful assistant with broad knowledge.
The provided context may or may not be accurate.

Carefully weigh the context against your own knowledge:
If the context seems accurate and relevant, you may rely on it.
If the context seems incomplete, irrelevant, or possibly incorrect, prefer your
own knowledge.
If neither the context nor your knowledge is sufficient, reply exactly
``NOT\_SUPPORTED''.

Do not explain or add extra words. Output only the final answer.

Question: \{question\}
Context: \{context\}
FinalAnswer:
\end{quote}

\subsection*{Mitigation Prompts}

\paragraph{Knowledge-Anchor.}
\begin{quote}\small\raggedright
You are a fact-checking assistant.

Step 1: Based on your own knowledge, determine the answer to this question first.
Step 2: Now read the provided context. Does it support or contradict your
knowledge-based answer?
Step 3:
  If the context supports your answer or adds useful detail $\to$ answer using that.
  If the context contradicts what you know $\to$ the context is likely wrong; stick
  with your knowledge.
  If you have no prior knowledge $\to$ use the context carefully; say NOT\_SUPPORTED
  if it is insufficient.

Do not explain or add extra words. Output only the final answer.

Question: \{question\}
Context: \{context\}
FinalAnswer:
\end{quote}

\paragraph{Adversarial-Aware.}
\begin{quote}\small\raggedright
You are a resilient question-answering assistant. The provided context may be
accurate, partially accurate, or deliberately misleading.

Priority rule: Your own knowledge takes precedence over the provided context.
If the context matches your knowledge: answer using either source.
If the context contradicts your knowledge: trust your knowledge; the context is
likely wrong.
If you have no prior knowledge on this topic: use the context cautiously.
If neither the context nor your knowledge is sufficient: reply exactly
``NOT\_SUPPORTED''.

Do not be misled by a context that sounds authoritative but contradicts facts
you know.
Do not explain or add extra words. Output only the final answer.

Question: \{question\}
Context: \{context\}
FinalAnswer:
\end{quote}

\section{Experimental Configuration}
\label{app:config}

\paragraph{Model configurations and knowledge cutoffs.}
All models were accessed via OpenRouter at temperature~0.
Table~\ref{tab:config} lists the evaluated models, providers, reported
knowledge cutoffs, and API identifiers. We refer to the API configuration \texttt{openai/o4-mini-high}
simply as o4-mini.

\begin{table*}[t]
\centering
\small
\setlength{\tabcolsep}{4pt}
\renewcommand{\arraystretch}{1.08}
\caption{Model configurations and API identifiers.}
\label{tab:requested_ids}\label{tab:config}
\begin{tabular}{@{}>{\raggedright\arraybackslash}p{3.9cm}p{2.4cm}p{1.4cm}p{5.8cm}@{}}
\toprule
Model / role & Provider & Cutoff & API identifier \\
\midrule
Generator / primary judge & \mbox{Google DeepMind} & --- & \texttt{google/gemini-3-flash} \\
DeepSeek-V3~\cite{deepseekai2024deepseekv3} & DeepSeek & Jul 2024 & \texttt{deepseek/deepseek-chat} \\
Llama-4-M~\cite{meta2025llama4} & Meta & Aug 2024 & \texttt{meta-llama/llama-4-maverick} \\
GPT-4.1-Mini~\cite{openai2025gpt41} & OpenAI & Jun 2024 & \texttt{openai/gpt-4.1-mini} \\
GPT-5.1~\cite{openai2026gpt51} & OpenAI & Sep 2024 & \texttt{openai/gpt-5.1} \\
o4-mini~\cite{openai2025o4mini} & OpenAI & Jun 2024 & \texttt{openai/o4-mini-high} \\
Claude-3.7~\cite{anthropic2025claude37} & Anthropic & Oct 2024 & \texttt{anthropic/claude-3-7-sonnet} \\
Independent judge & OpenAI & --- & \texttt{openai/gpt-5.1} \\
\bottomrule
\end{tabular}
\par\smallskip
{\footnotesize Claude-3.7 uses the October 2024 knowledge cutoff in its system card,
which separately states that public training data extend through November 2024
\cite{anthropic2025claude37}.}
\end{table*}

Exact routed snapshots and original access dates were not retained
(see Limitations).

\paragraph{Judge rubric.}
The following compact rubric defines the four labels. It is a summary
of the label rules, rather than a verbatim API message payload. The
additional-run judging scripts pass the question, prediction,
gold answer, and full passage to a shared context-aware rubric, with
Reject checked before Stick, Follow, and Wrong.
\begin{quote}\small\raggedright
Given a question and a model response, classify the response into exactly one
of the following labels:
Stick: the response is semantically equivalent to the gold answer.
Follow: the response is semantically equivalent to the false answer.
Reject: the model abstains or responds with NOT\_SUPPORTED.
Wrong: the response is incorrect and matches neither the gold nor the false answer.

Question: \{question\}
Gold answer: \{gold\}
False answer: \{false\}
Model response: \{response\}
Output only the label.
\end{quote}

We release a stratified 100-item passage-review sample.
Its review criteria cover support for the gold answer in true passages,
plausibility and gold-answer leakage in false passages, and answer
availability in insufficient passages.

\paragraph{Independent second-judge validation.}
Because the primary judge (Gemini-3-Flash) also generated the
false contexts, we validate its labels with an independent judge,
GPT-5.1, on a sample of 200 model responses stratified by primary-judge
label (50 each of Stick/Follow/Reject/Wrong), using the identical judge prompt.
The two judges agree on 92.0\% of items (Cohen's $\kappa = 0.893$).
Agreement by primary-judge label is Stick 100\%, Reject 98\%,
Follow 90\%, and Wrong 80\% (Table~\ref{tab:judge_confusion}).
Most disagreements involve responses originally labeled Wrong, with
further disagreements among Follow responses. Both validation cohorts
measure agreement between LLM judges; neither includes human adjudication.

\begin{table}[!htbp]
\centering
\footnotesize
\setlength{\tabcolsep}{5pt}
\caption{Judge agreement (200 responses).}
\label{tab:judge_confusion}
\begin{tabular}{@{}lrrrr@{}}
\toprule
Primary $\backslash$ 2nd & Stick & Follow & Reject & Wrong \\
\midrule
Stick  & 50 &  0 &  0 &  0 \\
Follow &  4 & 45 &  0 &  1 \\
Reject &  0 &  0 & 49 &  1 \\
Wrong  &  7 &  3 &  0 & 40 \\
\bottomrule
\end{tabular}
\end{table}

\paragraph{Four-prompt comparison validation.}
The four-prompt comparison adds seven model--prompt cells of
902 responses each, totaling 6,314 outputs: Passage-first and
Soft runs for DeepSeek-V3, Llama-4-M, and GPT-4.1-Mini, plus
Soft for GPT-5.1. The corresponding cells for o4-mini and GPT-5.1's
Passage-first cell were already available and are not included in
this new validation cohort. These outputs use the same 902 questions
and false passages; the subset covers December 2025--April 2026.

GPT-5.1 re-evaluates every new output. The independent-judge
script does not explicitly set sampling temperature; its output labels
are retained for reanalysis. The judges agree on 6,278/6,314
labels (99.4\%; $\kappa=0.943$). Among responses originally labeled
Follow, agreement is 5,969/5,997 (99.5\%). Adoption rates in the paper use the
primary judge's labels; independent labels are retained for the agreement
analysis.

\begin{table}[!htbp]
\centering
\footnotesize
\setlength{\tabcolsep}{5pt}
\caption{Judge agreement (6,314 responses).}
\label{tab:expanded_judge_confusion}
\begin{tabular}{@{}lrrrr@{}}
\toprule
Primary $\backslash$ second & Stick & Follow & Reject & Wrong \\
\midrule
Stick & 174 & 4 & 0 & 1 \\
Follow & 19 & 5969 & 0 & 9 \\
Reject & 1 & 0 & 128 & 0 \\
Wrong & 1 & 1 & 0 & 7 \\
\bottomrule
\end{tabular}
\end{table}

\begin{table}[!htbp]
\centering
\footnotesize
\setlength{\tabcolsep}{4pt}
\caption{Follow-label agreement by model.}
\label{tab:expanded_judge_models}
\begin{tabular}{@{}lrrr@{}}
\toprule
Model & All outputs & Follow agreement & \% \\
\midrule
DeepSeek-V3 & 1804 & 1720/1728 & 99.5 \\
Llama-4-M & 1804 & 1698/1702 & 99.8 \\
GPT-4.1-Mini & 1804 & 1719/1731 & 99.3 \\
GPT-5.1 & 902 & 832/836 & 99.5 \\
\bottomrule
\end{tabular}
\end{table}

The two validation cohorts cover different response sets and sampling
designs: the original sample balances primary labels, while the expanded
check covers every output in the seven new cells listed above.
Table~\ref{tab:expanded_judge_confusion} uses Gemini primary labels as
rows and independent GPT-5.1 labels as columns. In
Table~\ref{tab:expanded_judge_models}, each agreement denominator is the
number of primary Follow labels for that model.

Release contents and licensing are described in Appendix~\ref{app:pipeline}.

\section{Four-Prompt Comparison: Paired Results}
\label{app:expanded_ablation}

Table~\ref{tab:2x2} gives the four-condition adoption rates. Table~\ref{tab:paired_ablation} gives the discordant counts used in exact two-sided McNemar tests.
The discordant cells use the definitions in Appendix~\ref{app:stats};
the paired rate difference is $100(b-c)/902$ percentage points.
All comparisons here are exploratory.

Moving from Passage-first to Verify reduces adoption in all five models
(exact paired McNemar tests, all $p<0.001$).

\begin{table*}[t]
\centering
\small
\renewcommand{\arraystretch}{1.08}
\setlength{\tabcolsep}{5pt}
\caption{Paired prompt contrasts (902-item subset).}
\label{tab:paired_ablation}\label{tab:expanded_ctx}\label{tab:expanded_knowledge}\label{tab:expanded_verify}
\begin{tabular*}{\textwidth}{@{\extracolsep{\fill}}lrrrr@{}}
\toprule
Model & $b$ & $c$ & Change (pp) & Exact $p$ \\
\midrule
\multicolumn{5}{l}{\textit{Passage-first minus Baseline}} \\
DeepSeek-V3 & 22 & 5 & $+1.9$ & 0.002 \\
Llama-4-M & 54 & 0 & $+6.0$ & $<0.001$ \\
GPT-4.1-Mini & 22 & 1 & $+2.3$ & $<0.001$ \\
GPT-5.1 & 33 & 2 & $+3.4$ & $<0.001$ \\
o4-mini & 136 & 0 & $+15.1$ & $<0.001$ \\
\midrule
\multicolumn{5}{l}{\textit{Verify minus Passage-first}} \\
DeepSeek-V3 & 2 & 45 & $-4.8$ & $<0.001$ \\
Llama-4-M & 1 & 159 & $-17.5$ & $<0.001$ \\
GPT-4.1-Mini & 0 & 22 & $-2.4$ & $<0.001$ \\
GPT-5.1 & 2 & 40 & $-4.2$ & $<0.001$ \\
o4-mini & 2 & 18 & $-1.8$ & $<0.001$ \\
\midrule
\multicolumn{5}{l}{\textit{Soft minus Passage-first}} \\
DeepSeek-V3 & 9 & 31 & $-2.4$ & $<0.001$ \\
Llama-4-M & 1 & 43 & $-4.7$ & $<0.001$ \\
GPT-4.1-Mini & 2 & 29 & $-3.0$ & $<0.001$ \\
GPT-5.1 & 4 & 43 & $-4.3$ & $<0.001$ \\
o4-mini & 0 & 256 & $-28.4$ & $<0.001$ \\
\midrule
\multicolumn{5}{l}{\textit{Verify minus Baseline}} \\
DeepSeek-V3 & 10 & 36 & $-2.9$ & $<0.001$ \\
Llama-4-M & 12 & 116 & $-11.5$ & $<0.001$ \\
GPT-4.1-Mini & 9 & 10 & $-0.1$ & 1.000 \\
GPT-5.1 & 15 & 22 & $-0.8$ & 0.324 \\
o4-mini & 123 & 3 & $+13.3$ & $<0.001$ \\
Claude-3.7 & 29 & 133 & $-11.5$ & $<0.001$ \\
\bottomrule
\end{tabular*}
\end{table*}

\section{Statistical Testing Details}
\label{app:stats}

Rates and changes are rounded independently from their underlying values.
All prompt comparisons on the same item set are paired by item.  For
false-answer adoption rates, significance is assessed with McNemar tests on the
two discordant transition cells: $b$ for Baseline non-adoption to Verify adoption,
and $c$ for Baseline adoption to Verify non-adoption. Confidence intervals for
$\Delta$ are non-parametric bootstrap intervals over items with 5,000
resamples.  The cross-cohort Baseline comparison in Table~\ref{tab:shield} uses
two-proportion $z$-tests across temporal cohorts.  As stated in the main text,
the o4-mini post-cutoff McNemar comparison is the primary test; all other
significance markers are exploratory and are not corrected across the full
comparison family.

\section{Paired Transition Counts}
\label{app:transitions}

\paragraph{Pre-cutoff answer displacement.}
The pre-cutoff runs cover January 2020--April 2024 and contain
1,825 matched Baseline--Verify items per model and passage condition.
We verify identical questions, contexts, gold answers, and planted answers
within pairs. False-passage Stick-to-Follow counts are 171 for DeepSeek-V3,
152 for Llama-4-M, 176 for GPT-4.1-Mini, 552 for GPT-5.1, and 707
for o4-mini. Total transitions entering Follow are respectively 188,
166, 203, 599, and 976. On true passages, Stick-to-Reject accounts for
385 of 403, 321 of 348, 321 of 351, 536 of 550, and 398 of 425
transitions leaving Stick, in the same model order. These directional
counts describe displaced answers; net accuracy changes also include
responses entering Stick.

\paragraph{Recovery of the same displaced answers.}
Restricting to Baseline Stick and Verify Follow on pre-cutoff false
passages, Adversarial-Aware restores Stick on 141/171 items for
DeepSeek-V3, 144/152 for Llama-4-M, 163/176 for GPT-4.1-Mini,
546/552 for GPT-5.1, and 611/707 for o4-mini. Among items where Baseline
answers correctly with both passage types but Verify follows the false
passage and rejects the true passage, Adversarial-Aware restores Stick
on both versions in respectively 36/42, 33/37, 34/41, 184/186, and
169/195 cases. These comparisons track identical item identifiers
across the three prompts and both passage types.

\paragraph{Where reduced post-cutoff adoption goes.}
We merge the 900-item November 2024--November 2025 batch and the
902-item December 2025--April 2026 batch by item identifier. All six
models have 1,802 matched Baseline--Verify responses for both false
and true passages; questions, passages, gold answers, and planted
answers match within each pair. Among false-passage responses leaving
Follow, the destinations (Stick / Reject / Wrong) are 4 / 243 / 0 for
Claude-3.7, 3 / 95 / 0 for DeepSeek-V3, 3 / 239 / 0 for Llama-4-M,
3 / 18 / 0 for GPT-4.1-Mini, 6 / 43 / 0 for GPT-5.1, and 0 / 5 / 0
for o4-mini. On true passages, Stick-to-Reject counts are respectively
290, 351, 339, 73, 162, and 74. These are directional transition counts;
responses moving in the opposite direction also contribute to net rates.

For each model, the McNemar test compares two discordant cells on false
passages, using the discordant-cell definitions in
Appendix~\ref{app:stats}.

Table~\ref{tab:transitions_post} shows the full $2\times 2$ paired matrix for
o4-mini on both temporal cohorts, and Table~\ref{tab:transitions_all}
summarizes $b$ and $c$ for all models.  The pre-cutoff matrix has 1,825 paired items
because two items lack paired Baseline and Verify labels.

\begin{table*}[t]
\centering
\renewcommand{\arraystretch}{1.12}
\small
\setlength{\tabcolsep}{4pt}
\caption{False-passage discordant counts.}
\label{tab:transitions_all}
\begin{tabular*}{\textwidth}{@{\extracolsep{\fill}}lrrrrrr@{}}
\toprule
 & \multicolumn{3}{c}{Pre-cutoff} & \multicolumn{3}{c}{Post-cutoff} \\
\cmidrule(lr){2-4}\cmidrule(lr){5-7}
Model & $b$ & $c$ & ratio & $b$ & $c$ & ratio \\
\midrule
DeepSeek-V3 & 188 & 62 & 3.0 & 19 & 98 & 0.2 \\
Llama-4-M & 166 & 56 & 3.0 & 26 & 242 & 0.1 \\
GPT-4.1-Mini & 203 & 29 & 7.0 & 19 & 21 & 0.9 \\
GPT-5.1 & 599 & 13 & 46.1 & 36 & 49 & 0.7 \\
o4-mini & 976 & 6 & 162.7 & 251 & 5 & 50.2 \\
Claude-3.7 & --- & --- & --- & 55 & 247 & 0.2 \\
\bottomrule
\end{tabular*}
{\small Ratio $= b/c$.  Claude-3.7 was unavailable for pre-cutoff.}
\end{table*}

\section{Case Studies: Override of Correct No-Context Answers}
\label{app:case_studies}

The pre-cutoff example in Figure~\ref{fig:phenomenon} concerns Biden's
HHS nominee (item \texttt{US:\allowbreak2020-12:\allowbreak0080}).
The gold answer is Xavier Becerra and the planted answer is Julian Castro;
o4-mini answers correctly under No context and Baseline, then gives
Castro under Verify.

Table~\ref{tab:case_studies} shows three illustrative items from the
post-cutoff cohort where o4-mini replaces a correct no-context answer.
Correct no-context answers can reflect earlier background facts or
information inferable from the question; these cases establish answer
displacement, independently of when the fact was learned. The Windows 10
example is item \texttt{World:2025-10:0327}; dates are shortened in the table.
The Fairfax item belongs to an April 2026 TiEBe entry but asks about
his 2018--2022 office. Cohort membership follows the source-event
timestamp, while the question can concern an earlier background fact.
In each case the model answers correctly under No context and Baseline, but switches to the false answer under Verify.
Gold is the correct answer, False is
the answer planted in $C_F$, and No context/Baseline/Verify are the model responses under
each prompt.

\begin{table*}[!htbp]
\centering
\footnotesize
\setlength{\tabcolsep}{4pt}
\caption{Displacement of correct no-context answers.}
\label{tab:case_studies}
\begin{tabular}{@{}>{\raggedright\arraybackslash}p{4.7cm}
>{\raggedright\arraybackslash}p{1.9cm}
>{\raggedright\arraybackslash}p{1.9cm}
>{\raggedright\arraybackslash}p{1.8cm}
>{\raggedright\arraybackslash}p{1.8cm}
>{\raggedright\arraybackslash}p{1.8cm}@{}}
\toprule
Question & Gold & False (in context) & No context & Baseline & Verify \\
\midrule
How long had the insurgency waged by Boko Haram in Borno State lasted by April 2026?
  & 17 years & 15 years
  & 17 years & 17 years & 15 years \\[4pt]
What political office did Justin Fairfax hold in Virginia between 2018 and 2022?
  & Lieutenant Governor & Attorney General
  & Lt.\ Gov.\ of Virginia & Lt.\ Gov.\ of Virginia & Attorney General \\[4pt]
On what date did Microsoft end technical support and security updates for Windows 10?
  & Oct. 14, 2025 & Oct. 10, 2028
  & Oct. 14, 2025 & Oct. 14, 2025 & Oct. 10, 2028 \\
\bottomrule
\end{tabular}
\par\smallskip
{\footnotesize Verify responses show false-answer adoption.}
\end{table*}


\paragraph{Follow-up explanation probe.}
For the 31 selected post-cutoff cases with a correct no-context answer,
a correct Baseline answer, and a false Verify answer, we
re-queried o4-mini with an explanation prompt. It asked the model to
state what it knew, assess agreement with the passage, and then answer.
The model adopted the false answer in 27 of 31 cases. Its
responses did not mention the prior answer or flag a conflict; longer
responses sometimes argued for the false claim.

\section{Additional Results by Passage Type}
\label{app:full_results}

Table~\ref{tab:reference_outcomes} reports the reference outcomes:
Baseline abstention on insufficient passages in both cohorts,
context-only false-answer adoption, and no-context accuracy on post-cutoff
items. Core Baseline/Verify rates are in Table~\ref{tab:core_results}.
Positive $\Delta$ values indicate increased adoption under Verify.
Bootstrap 95\% intervals for the post-cutoff $\Delta$ values are:
DeepSeek-V3 [$-5.3,-3.5$], Llama-4-M [$-13.3,-10.7$],
GPT-4.1-Mini [$-0.6,+0.4$], GPT-5.1 [$-1.5,+0.1$], o4-mini
[$+12.4,+14.9$], and Claude-3.7 [$-12.0,-9.2$].

\begin{table*}[t]
\centering
\small
\renewcommand{\arraystretch}{1.12}
\setlength{\tabcolsep}{5pt}
\caption{Reference outcomes by temporal cohort.}
\label{tab:reference_outcomes}\label{tab:full_postcutoff}\label{tab:full_precutoff}
\begin{tabular*}{\textwidth}{@{\extracolsep{\fill}}lrrrr@{}}
\toprule
 & \multicolumn{2}{c}{Baseline abstention on insufficient passages} & \multicolumn{2}{c}{Post-cutoff references} \\
\cmidrule(lr){2-3}\cmidrule(lr){4-5}
Model & Pre-cutoff & Post-cutoff & \shortstack{Context only\\false adoption} & \shortstack{No-context\\accuracy} \\
\midrule
DeepSeek-V3 & 31.6 & 62.4 & 92.6 & 1.3 \\
Llama-4-M & 14.7 & 64.4 & 91.0 & 1.8 \\
GPT-4.1-Mini & 6.1 & 27.0 & 95.0 & 4.8 \\
GPT-5.1 & 5.3 & 59.7 & 92.8 & 4.6 \\
o4-mini & 7.8 & 47.0 & 94.8 & 10.0 \\
Claude-3.7 & --- & 80.9 & 88.5 & 1.8 \\
\bottomrule
\end{tabular*}
\par\smallskip
{\footnotesize Rates are percentages. Claude-3.7 was unavailable for the pre-cutoff evaluation.}
\end{table*}

\section{Supplementary Result Tables}
\label{app:supp}

These tables support four analyses summarized in the main text:
the cross-cohort Baseline contrast (Table~\ref{tab:shield}), the Verify prompt-wording
ablation (Table~\ref{tab:ablation}), the o4-mini mitigation on the
902-item subset (Table~\ref{tab:mitigation}), and the Baseline/mitigation group rates
(Table~\ref{tab:mitigation_strat}).
$\Delta_S$ is the pre-cutoff minus post-cutoff Baseline adoption rate,
tested with a two-proportion $z$-test. In Table~\ref{tab:ablation}, all
variants differ from Verify by McNemar test ($p{<}0.001$).  In
Table~\ref{tab:mitigation}, the true-passage accuracy change is relative
to Baseline. Rates are percentages and changes are percentage points.

\begin{table*}[t]
\centering
\small
\renewcommand{\arraystretch}{1.12}
\setlength{\tabcolsep}{5pt}
\caption{Baseline adoption by temporal cohort.}
\label{tab:shield}
\begin{tabular*}{\textwidth}{@{\extracolsep{\fill}}lrrr@{}}
\toprule
 & \multicolumn{3}{c}{Baseline adoption (\%)} \\
\cmidrule(lr){2-4}
Model & Pre-cutoff & Post-cutoff & Change ($\Delta_S$) \\
\midrule
DeepSeek-V3 & 79.9 & 93.4 & $-13.5^{***}$ \\
Llama-4-M & 80.8 & 88.8 & $-8.0^{***}$ \\
GPT-4.1-Mini & 82.1 & 94.1 & $-12.0^{***}$ \\
GPT-5.1 & 42.7 & 92.0 & $-49.3^{***}$ \\
o4-mini & 33.9 & 80.5 & $-46.7^{***}$ \\
Claude-3.7 & --- & 71.6 & --- \\
\bottomrule
\end{tabular*}
\end{table*}

\begin{table*}[t]
\begin{minipage}[t]{0.48\textwidth}
\centering
\footnotesize
\renewcommand{\arraystretch}{1.12}
\setlength{\tabcolsep}{4pt}
\caption{Prompt controls (o4-mini; 902-item subset).}
\label{tab:ablation}
\begin{tabular*}{\linewidth}{@{\extracolsep{\fill}}lr@{}}
\toprule
Prompt & Adoption (\%) \\
\midrule
Baseline & 82.3 \\
Context only & 96.6 \\
Verify & 95.6 \\
Brief & 89.5 \\
CoT-Extended & 93.0 \\
Soft & 69.0 \\
\bottomrule
\end{tabular*}
\end{minipage}
\hfill
\begin{minipage}[t]{0.48\textwidth}
\centering
\footnotesize
\renewcommand{\arraystretch}{1.12}
\setlength{\tabcolsep}{3pt}
\caption{Mitigation results (o4-mini; 902-item subset).}
\label{tab:mitigation}
\begin{tabular*}{\linewidth}{@{\extracolsep{\fill}}l@{\hspace{8pt}}r@{\hspace{8pt}}r@{}}
\toprule
Prompt & \shortstack{Adoption\\(\%)} & \shortstack{Accuracy\\change (pp)} \\
\midrule
Baseline & 82.3 & --- \\
Verify & 95.6 & $-2.5^{**}$ \\
Knowledge-Anchor & 84.1 & $+1.1$ (ns) \\
Adversarial-Aware & 75.2 & $-2.5^{**}$ \\
\bottomrule
\end{tabular*}
\par\smallskip
{\footnotesize $^{**}$ denotes $p<0.01$; ns denotes not significant.}
\end{minipage}
\end{table*}

\begin{table*}[t]
\centering
\small
\renewcommand{\arraystretch}{1.12}
\setlength{\tabcolsep}{4pt}
\caption{Pre-cutoff mitigation by known/unknown items.}
\label{tab:mitigation_strat}
\begin{tabular*}{\textwidth}{@{\extracolsep{\fill}}lrrrrrr@{}}
\toprule
 & \multicolumn{3}{c}{Known} & \multicolumn{3}{c}{Unknown} \\
\cmidrule(lr){2-4}\cmidrule(lr){5-7}
Model & Baseline & Adv.-Aware & Change (pp) & Baseline & Adv.-Aware & Change (pp) \\
\midrule
DeepSeek-V3 & 63.2 & 29.8 & $-33.4$ & 91.6 & 68.6 & $-23.0$ \\
Llama-4-M & 70.0 & 47.8 & $-22.2$ & 92.8 & 82.1 & $-10.7$ \\
GPT-4.1-Mini & 74.1 & 53.5 & $-20.5$ & 94.0 & 84.5 & $-9.5$ \\
GPT-5.1 & 33.1 & 12.9 & $-20.3$ & 76.4 & 54.1 & $-22.3$ \\
o4-mini & 21.9 & 13.8 & $-8.1$ & 53.8 & 44.0 & $-9.8$ \\
\bottomrule
\end{tabular*}
\par\smallskip
{\footnotesize Rates are percentages; negative changes indicate reduced adoption.}
\end{table*}

\end{document}